\pdfoutput=1
\documentclass[11pt]{article}

\usepackage{EMNLP2022}

\usepackage{times}
\usepackage{latexsym}

\usepackage[T1]{fontenc}
\usepackage[utf8]{inputenc}

\usepackage{microtype}

\usepackage{inconsolata}

\newcommand{\IC}{I\textsc{mage}C\textsc{o}D\textsc{e}}
\usepackage{url}
\usepackage{paralist}
\usepackage{diagbox}

\usepackage{graphicx}
\usepackage{subfig}
\usepackage{xcolor, soul}

\title{Communication breakdown: On the low mutual intelligibility between human and neural captioning}

\author{Roberto Dess\`{i} \\
  Meta AI \\
  Universitat Pompeu Fabra \\
  \texttt{rdessi@meta.com} \\\And
  {\bf Eleonora Gualdoni} \\\and
  {\bf Francesca Franzon} \\
  Universitat Pompeu Fabra \\
  \texttt{\{name.lastname\}@upf.edu} \\\And
  {\bf Gemma Boleda} \\\and
  {\bf Marco Baroni} \\
  ICREA \\
  Universitat Pompeu Fabra \\
  \texttt{\{name.lastname\}@upf.edu} \\
}

\begin{document}
\maketitle
\begin{abstract}
We compare the 0-shot performance of a neural caption-based image retriever when given as input either human-produced captions or captions generated by a neural captioner. We conduct this comparison on the recently introduced \IC{} data-set \citep{Krojer:etal:2022}, which contains hard distractors nearly identical to the images to be retrieved. We find that the neural retriever has much higher performance when fed neural rather than human captions, despite the fact that the former, unlike the latter, were generated without awareness of the distractors that make the task hard. Even more remarkably, when the same neural captions are given to human subjects, their retrieval performance is almost at chance level. Our results thus add to the growing body of evidence that, even when the ``language'' of neural models resembles English, this superficial resemblance might be deeply misleading.\end{abstract}

\section{Introduction}

Neural vision-and-language models have achieved impressive results in tasks such as visual commonsense reasoning and question answering \citep[e.g.,][]{Chen:etal:2020,Lu:etal:2019}. However, \citet{Krojer:etal:2022} recently showed, in the context of caption-based image retrieval, that state-of-the-art multimodal models still perform poorly when the candidate  pool contains very similar distractor images (such as close frames from the same video). 
 
Here, we show that, when the best pre-trained image retrieval system of \citet{Krojer:etal:2022} is fed captions produced by an out-of-the box neural caption generator, its performance makes a big jump forward. 0-shot  image retrieval accuracy improves by almost 6\% compared to the highest previously reported human-caption-based performance by the same model, with fine-tuning and various \textit{ad-hoc} architectural adaptations. This is remarkable, because the off-the-shelf caption generator we use (unlike the humans who wrote the original captions in the data-set) is \textit{not} taking the set of distractor images into account. Even more remarkably, we show that, when human subjects are tasked with retrieving the right image using the same neural captions that help the model so much, their performance is only marginally above chance level.

\section{Setup}

\paragraph{Data} We use the more challenging \textit{video} section of the \IC{} data-set \citep{Krojer:etal:2022}. Since we do not fine-tune our model, we only use the validation set, including 1,872 data points.\footnote{We use the validation set because \IC{} test set annotations are not publicly available.} Henceforth, when when we employ the term \IC{}, we are referring to this subset. Each data-point consists of a target image and 9 distractors, where the target and the distractors are frames from the same (automatically segmented) scene in a video. We also use the human captions in the data-set, that were produced by subjects that had access to the distractors while annotating each target (they were instructed to take distractors into account, without explicitly referring to them). Having access to this ``common ground'' \citep{Brennan:Clark:1996}, annotators produced highly context-dependent descriptions (see example human captions in Fig.~\ref{fig:SixExamples}). The data-set contains one single caption per image.

\paragraph{Neural caption generation} We use the ClipCap caption generation system \citep{Mokady:etal:2021} without fine-tuning. For details and hyperparameters of the generation process see Appendix \ref{appendix:model-details}. In short, ClipCap processes an image with a CLIP visual encoder \citep{Radford:etal:2021} and learns a mapping from the resulting visual embedding to a sequence of embeddings in GPT-2 space \citep{Radford:etal:2019}, that are used to kickstart the generation of a sequence of tokens. We report experiments with the ClipCap variant fine-tuned on the COCO data-set \cite{Lin:etal:2014}, where the weights of the multimodal mapper were updated and those of the language model (GPT-2) were kept frozen. We obtained very similar results with the other publicly available ClipCap variants. We generate a single \textit{neural caption} for each \IC{} target image by passing it through ClipCap. Note that, as there is no way to make this out-of-the-box architecture distractor-aware, the neural captions do not take distractors into account.

% \paragraph{Image retrieval} We use the simplest CLIP-based retrieval system of \citet{Krojer:etal:2022} (lacking their ``context module'', temporal embeddings and using a ViT-B/16 backbone \cite{Dosovitskiy:etal:2021}). 
% %The caption is passed through the CLIP text encoder. Each image in a set is passed through the CLIP visual encoder. 
% The caption and each image in the set are passed through a Transformer-based text encoder and a Transformer-based visual encoder, respectively.
% Retrieval is successful if the dot product between the caption representation and the image representation of the target is larger than that of the caption with any other distractor. The ClipCap pretrained models use a ViT-B/32  backbone \cite{Dosovitskiy:etal:2021}, whereas \citet{Krojer:etal:2022} used the smaller ViT-B/16 architecture. Thus, for fair comparison, we re-computed 0-shot retrieval accuracies on \IC{} using a ViT-B/32 model.
\paragraph{Image retrieval} We use the simplest architecture proposed by \citet{Krojer:etal:2022} (the one without context module and temporal embeddings), which amounts to a standard CLIP retriever from \citet{Radford:etal:2021}. The caption and each image in the set are passed through textual and visual encoders, respectively. Retrieval is successful if the dot product between the resulting caption and target image representations is larger than that of the embedded caption with any distractor representation. We use the ResNet-based CLIP visual encoder \cite{He:etal:2015}, whereas \citet{Krojer:etal:2022} used the ViT-B/16 architecture. We found the former to have a slightly higher 0-shot retrieval accuracy compared to the one they used (17.4\% in Table \ref{tab:acc-results} here vs.~14.9\% in their paper).

\section{Results and analysis}

\begin{figure*}[t]
    \centering
    \includegraphics[width = \hsize]{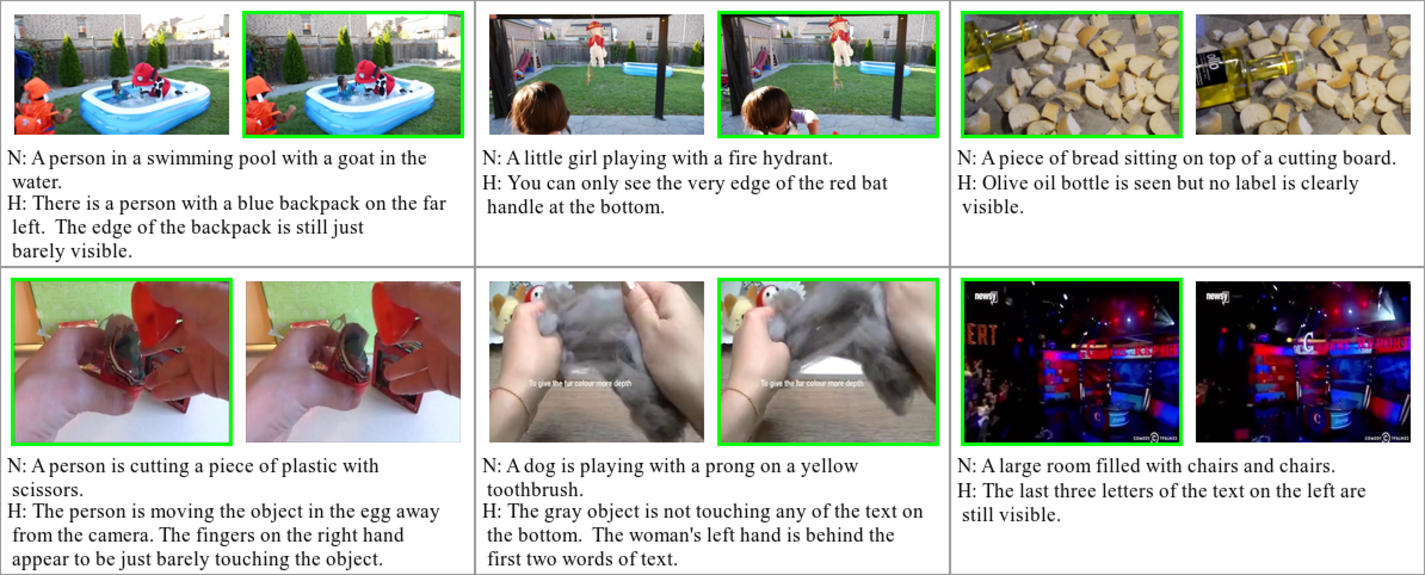}
    \caption{Randomly selected data points from \IC{} with neural (N) and human (H) captions, where, given the neural caption, the neural retriever guessed the target and the human retriever failed. For each candidate set, we show the target (marked in green) and (randomly) either the immediately preceding or following distractor frame.  Appendix \ref{appendix:example-figure-construction} reports the whole candidate sets for each data-point (10 images per set), as well as providing details on the random selection process.}
    \label{fig:SixExamples}
\end{figure*}

\begin{table}[tb]
\centering
\begin{tabular}{lc}
\hline
setup & acc\\
\hline
neural captions, 0-shot&27.9\\
human captions, 0-shot&17.4\\
human captions, Krojer et al's best&22.3\\
\hline
\end{tabular}
\caption{\label{tab:acc-results}
Percentage \IC{} accuracy of 0-shot image retriever when given neural vs.~human captions as input. Last row reports accuracy of best fine-tuned, architecturally-adjusted model from \citet{Krojer:etal:2022} (featuring a context module, temporal embeddings and a ViT-B/16 backbone).}
\end{table}

\paragraph{Neural vs.~human caption performance}  As shown in Table \ref{tab:acc-results}, the out-of-the-box neural image retrieval model has a clear preference for neural captions. It reaches 27.9\% \IC{} accuracy when taking neural captions as input, vs.~17.4\% with human captions (chance level is at 10\%).
For comparison, the best fine-tuned, architecture-adjusted model of \citet{Krojer:etal:2022} reached 22.3\% performance with human captions.

A concrete sense of the differences between the two types of captions is given by the examples in Fig.~\ref{fig:SixExamples}. The examples in this figure are  picked randomly. Based on manual inspection of a larger set, we are confident they are representative of the full data. Clearly, neural captions are shorter (avg.~length at 11.4 tokens vs.~23.2 for human captions) and more plainly descriptive (although the description is mostly only vaguely related to what's actually depicted). Since there is no way to make the out-of-the-box ClipCap system distractor-aware, the neural captions are not highlighting discriminative aspects of a target image compared to the distractors. Human captions, on the other hand, use very articulated language to highlight what is unique about the target compared to the closest distractors (often focusing on rather marginal aspects of the image, because of their discriminativeness, e.g., for the first example in the figure, the fact that the blue backpack is hardly visible). It is not surprising that a generic image retriever, that was not trained to handle this highly context-based linguistic style, would not get much useful information out of the human captions. It is interesting, however, that this generic system performs relatively well with the neural captions, given how off-the-mark and non-discriminative the latter typically are.

As more quantitative cues of the differences between caption types, we observe that human captions are making more use of both rare lemmas and function words (see frequency plots in Appendix \ref{appendix:frequency-analysis}).\footnote{Code to reproduce our analysis with human and model-generated captions is available at \url{https://github.com/franfranz/emecomm_context}}  Extracting the lemmas that are statistically most strongly associated to the human caption set (see Appendix \ref{appendix:typical-words} for method and full top list), we observe ``meta-visual'' words such as \textit{visible} and \textit{see}, pronouns and determiners cuing anaphoric structure (\textit{the}, \textit{her}, \textit{his}), and function words signaling a more complex sentence structure, such as auxiliaries, negation and connectives. Among the most typical neural lemmas, we find instead general terms for concrete entities such as \textit{people}, \textit{woman}, \textit{table} and \textit{food}.

\paragraph{Are neural captions really discriminative?}

By looking at Figure \ref{fig:SixExamples}, we see that neural captions might be (very noisily) descriptive of the target, but they seem hardly discriminative with respect to the nearest distractors. Recall that each \IC{} set contains a sequence of 10 frames from the same scene. In general, the frames that are farther away in time might be easier to discriminate than the closest ones (consider the full candidate set examples in Appendix \ref{appendix:example-figure-construction}: it is in general much easier to tell apart the first and last frames than two adjacent images). It could be, then, that the non-random but still low performance of the image retriever with neural captions is due to the combination of two factors. On the one hand, the neural captions might suffice for the retriever to exclude the farthest distractors. On the other, its performance at telling the closest frames apart is actually random.

To rule out this explanation, we repeated the retrieval experiment in the most challenging setup, in which we excluded all but the distractors immediately preceding and following the target frame (if the target is the first/last image, we pick the two frames following/preceding it, respectively). The retriever using neural captions still reaches 48.7\% accuracy, well above chance level (33.3\%). We must thus conclude that neural captions such as those in Fig.~\ref{fig:SixExamples} do carry a non-negligible degree of discriminative power for a neural image retriever.

On a related point, neural captions such as those in the figure seem so generic that one could imagine the neural caption generator would produce the same caption for close-by frames. This is not the case. We consider the 678 \IC{} cases in which a candidate set is repeated across multiple data points, with only the target changing, and in which the targets are adjacent or one-frame apart. In 93.7\% of such cases, ClipCap generated at least two non-identical captions. For example, the frame on the left of the center-bottom pair in Fig.~\ref{fig:SixExamples} is also used as a target, and for it ClipCap generates the caption ``A person holding a water bottle with a dog in it.'' Looking at the right-bottom pair, the frame on the right is also a target, and ClipCap generated the following caption for it: ``A picture of a kitchen with a bunch of televisions on it.'' Future research should ascertain to what extent these intuitively uninformative variations in frame description actually contain cues that are systematically discriminative for the retrieval system. 

 \paragraph{Human performance on neural captions} Looking again at examples such as those in Fig.~\ref{fig:SixExamples} (all cases in which the image retriever correctly identified the target), we might conjecture that the neural captions are more informative for the neural retriever model than they are for humans. To verify this hypothesis, we organized a crowd-sourcing experiment in which human subjects had to perform the same 10-image-set target discrimination task we submitted to the neural image retriever.

More precisely, we selected a subset of \IC{} that is balanced in terms of image retriever performance as follows. We used all 522 sets where the retriever guessed the target, and we randomly added the same number of sets where the retriever got it wrong (so that its accuracy, on this subset, is at 50\% by construction). We collected human discrimination decisions for this subset of 1,044 items, when either human or neural captions are given as input. We collected one rating per item-caption combination from a total of 36 Amazon Mechanical Turk\footnote{\url{https://www.mturk.com/}} participants, that each provided a total of 58 ratings. Experimental details are given in Appendix \ref{appendix:crowdsourcing}.

On the relevant \IC{} subset, humans clearly outperform the neural retriever when human captions are given: 54.3\% human discrimination accuracy vs.~16.3\% for the neural retriever.\footnote{See Appendix \ref{appendix:crowdsourcing} for discussions of why our human-to-human discrimination accuracy is considerably lower than that reported by \citet{Krojer:etal:2022} on the whole \IC{} data-set.} Strikingly, the pattern sharply reverses with neural captions: 50.0\% for the neural image retriever vs.~12.8\% for humans (not much above the 10\% random baseline).

We thus confirm that neural captions such as those presented in Fig.~\ref{fig:SixExamples}, despite being apparently vague and inaccurate descriptions of the target image ``in plain English'', carry significantly more discriminative value for the neural retriever than they do for humans (recall that the examples in this figure were selected among the cases where the neural caption allowed the neural retriever to guess the right target, while human subjects failed the task).

\begin{table}[tb]
\centering
\begin{tabular}{lcc}
\hline
\diagbox{retriever}{captions}&human & neural\\
\hline
human &54.3&12.8\\
neural&16.3&50.0\\
\hline
\end{tabular}
\caption{\label{tab:neural-human-acc-results}
Percentage accuracy on an \IC{} subset (balanced to get 50\% accuracy of the 0-shot neural retriever with neural captions): human vs.~neural retrievers tested with neural vs.~human captions as inputs.}
\end{table}

\section{Conclusion}

Previous research has shown that neural caption generators occasionally produce highly counterintuitive or irrelevant image descriptions \citep[e.g.,][]{Lake:etal:2016,Rohrbach:etal:2018}. We provide here evidence that such descriptions might only be misleading or uninformative for humans, while still being relatively ``understandable'' to neural models. We discuss below the \textbf{Limitations} that delimit the scope of our finding. Still, we can tentatively conclude that, even when they are trained on English, deep nets might pack and retrieve information from token sequences that are different from those an English speaker would encode in and extract from them.

A better understanding of this behaviour could help design higher-performance systems. For example, we could implement a module translating human captions to the ``machine code'' that neural models prefer, leading to better caption-based retrieval; or, from a model-to-model communication perspective \citep{Zeng:etal:2022}, optimize caption generation directly for neural model understanding, instead of imitating human captions. Similar ideas have recently proposed in the context of textual information retrieval \citep[e.g.,][]{Haviv:etal:2021,Shin:etal:2020}.

From a less optimistic perspective, our results can be interpreted as another cautionary tale about the degree to which neural models truly ``understand language'' \citep{Webson:Pavlick:2022}, and suggest that a good grasp of their counter-intuitive behaviour should be a priority of current research, or else malicious agents could rely on the models' opaque behaviour for adversarial attacks \citep{Wallace:etal:2019} and other types of model misuse.

\section*{Limitations}

The results we presented are limited to one specific data-set tested with a single caption generator and image retriever pair (with both systems relying on the CLIP image encoder). Future work should verify whether they generalize to other neural model combinations and data-sets.

We observe a considerable increase in accuracy when the neural image retriever is fed machine-generated captions instead of human ones. However, accuracy is still at 27.9\%, suggesting that the retrieval system has only a very partial understanding of captions, whether machine- or human-produced. How to improve its performance remains a question for future work. In the current setup, the caption generation system, unlike human annotators, only receives the target image as input, and it is unaware of the distractors. Making the caption generation system distractor-aware \citep[perhaps taking inspiration from work on ``image difference captioning'', e.g.,][]{Guo:etal:2022} might improve the performance of the image retriever. Distractor-aware neural captions would also be more fairly comparable to the distractor-aware human captions we got from the \IC{} data-set.

Last but not least, we provided evidence that captions that carry virtually no discriminative information for humans are instead helping the neural retriever identify target images well above chance level. We still lack, however, an understanding of \textit{how} the neural retriever accomplishes this surprising feat: developing such an understanding is perhaps the most important direction for future work.

\section*{Ethics Statement}

We rely on existing, publicly available data-sets \citep{Das:etal:2013,Krojer:etal:2022,Li:etal:2019,Xu:etal:2016, Lin:etal:2014}
and pre-trained models \citep{Mokady:etal:2021,Radford:etal:2021}.

We re-normed a subset of the data used by \citet{Krojer:etal:2022} using crowdsourcing. The experiment was approved by the ethical board of Universitat Pompeu Fabra in the context of the AMORE project (grant agreement No.\ 715154). Participants had to agree to an informed consent form before doing the experiment, and they were allowed to leave it at any time. No personal data were collected, except for the participants' AMT worker IDs, needed for their payment. They were paid 12.5\$ for completing the task (that took about 20 minutes). The crowdsourcing experiment procedure is described in more detail in Appendix \ref{appendix:crowdsourcing}. 

As we only run zero-shot experiments with pre-trained models, compute usage is negligible.

Our research contributes to an expanding body of evidence showing that, while pre-trained deep models are apparently responding to natural language prompts, their ``language'' might differ from human language \citep[e.g.,][]{Lu:etal:2022,Shin:etal:2020,Webson:Pavlick:2022}. Understanding this gap between human and machine language is important, in order to improve human-machine interaction, but also because it can be exploited for harmful purposes, such as adversarial attacks \citep{Wallace:etal:2019}.

\section*{Acknowledgements}

We would like to thank Michele Bevilacqua for comments on an earlier draft and the FAIR conference participants for fruitful discussion. We thank the EMNLP area chair and anonymous reviewers for feedback.

This project has received funding from the European Research Council (ERC) under the European Union's Horizon 2020 research and innovation programme (grant agreements No.\ 715154 and No.\ 101019291) and the Spanish Research Agency (ref.\ PID2020-112602GB-I00). This paper reflects the authors' view only, and the funding agencies are not responsible for any use that may be made of the information it contains.

\begin{flushright}
\includegraphics[width=0.8cm]{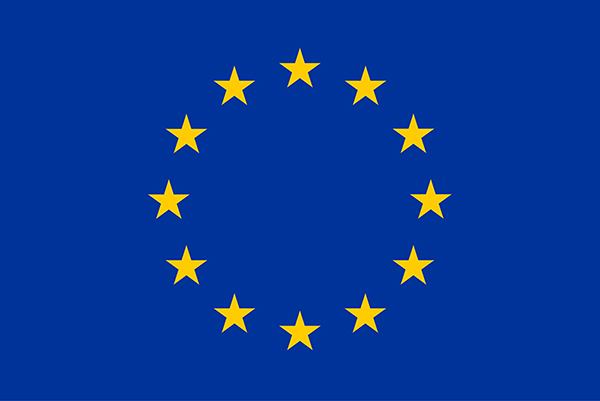}
\includegraphics[width=0.8cm]{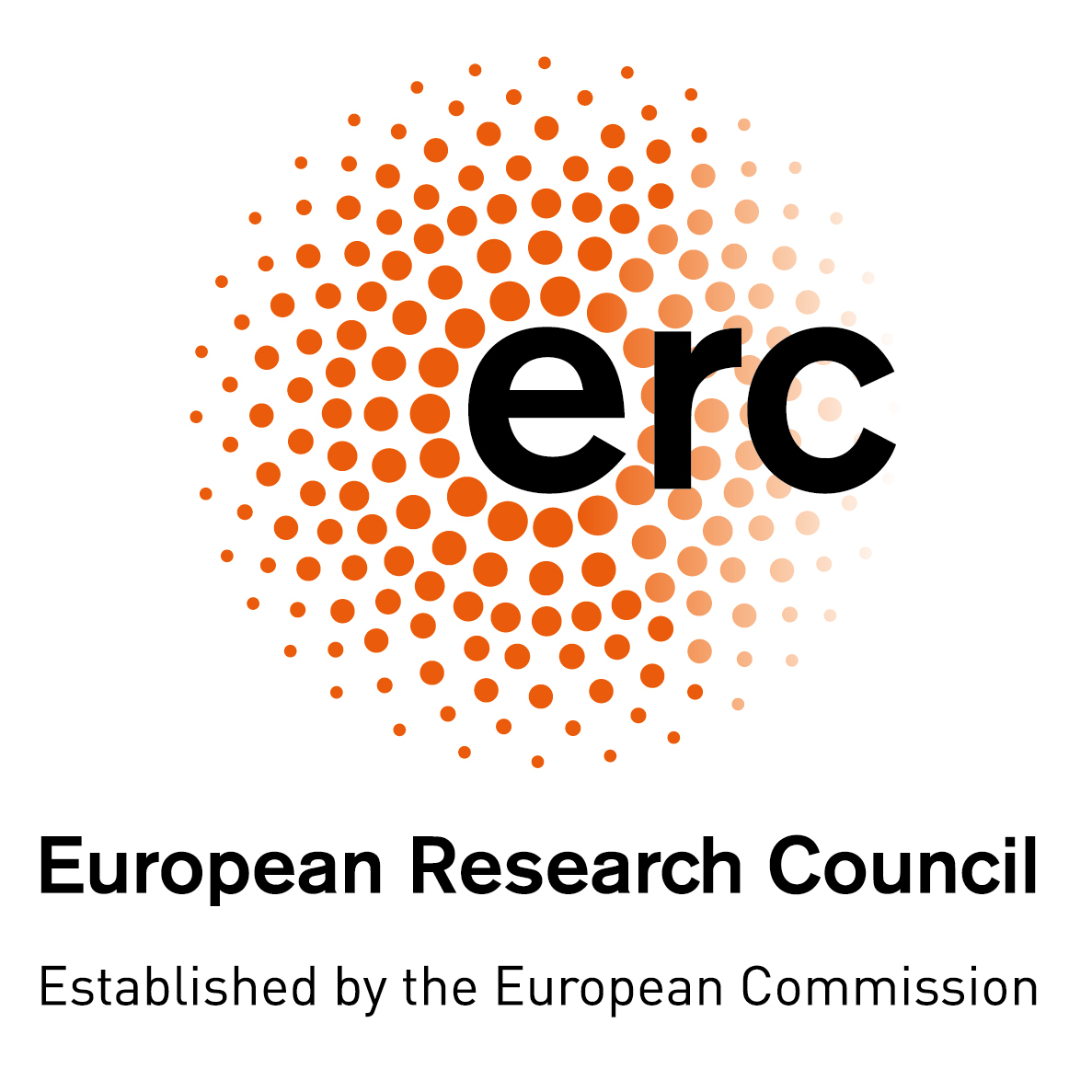}
\end{flushright}

% Entries for the entire Anthology, followed by custom entries
\bibliography{marco, others}

\begin{thebibliography}{24}
\expandafter\ifx\csname natexlab\endcsname\relax\def\natexlab#1{#1}\fi

\bibitem[{Baayen(2001)}]{Baayen:2001}
Harald Baayen. 2001.
\newblock \emph{Word Frequency Distributions}.
\newblock Kluwer, Dordrecht, The Netherlands.

\bibitem[{Brennan and Clark(1996)}]{Brennan:Clark:1996}
Susan Brennan and Herbert Clark. 1996.
\newblock Conceptual pacts and lexical choice in conversation.
\newblock \emph{Journal of Experimental Psychology: Learning, Memory, and
  Cognition}, 22(6):1482--1493.

\bibitem[{Chen et~al.(2019)Chen, Li, Yu, {El Kholy}, Ahmed, Gan, Cheng, and
  Liu}]{Chen:etal:2020}
Yen-Chun Chen, Linjie Li, Licheng Yu, Ahmed {El Kholy}, Faisal Ahmed, Zhe Gan,
  Yu~Cheng, and Jingjing Liu. 2019.
\newblock {UNITER}: {UNiversal Image-TExt Representation} learning.
\newblock In \emph{Proceedings of ECCV}, pages 104--120, virtual conference.

\bibitem[{Das et~al.(2013)Das, Xu, Doell, and Corso}]{Das:etal:2013}
Pradipto Das, Chenliang Xu, Richard Doell, and Jason Corso. 2013.
\newblock A thousand frames in just a few words: {Lingual} description of
  videos through latent topics and sparse object stitching.
\newblock In \emph{Proceedings of CVPR}, pages 2634--2641, Portland, Oregon.

\bibitem[{Evert(2005)}]{Evert:2005}
Stefan Evert. 2005.
\newblock \emph{The Statistics of Word Cooccurrences}.
\newblock Ph.{D} dissertation, Stuttgart University.

\bibitem[{Guo et~al.(2022)Guo, Tzu-Jui, Wang, and Laaksonen}]{Guo:etal:2022}
Zixin Guo, Tzu-Jui, Julius Wang, and Jorma Laaksonen. 2022.
\newblock {CLIP4IDC}: {CLIP} for image difference captioning.
\newblock \url{https://arxiv.org/abs/2206.00629}.

\bibitem[{Haviv et~al.(2021)Haviv, Berant, and Globerson}]{Haviv:etal:2021}
Adi Haviv, Jonathan Berant, and Amir Globerson. 2021.
\newblock {BERT}ese: Learning to speak to {BERT}.
\newblock In \emph{Proceedings of EACL}, pages 3618--3623, Online.

\bibitem[{He et~al.(2015)He, Zhang, Ren, and Sun}]{He:etal:2015}
Kaiming He, Xiangyu Zhang, Shaoqing Ren, and Jian Sun. 2015.
\newblock \href {https://doi.org/10.48550/ARXIV.1512.03385} {Deep residual
  learning for image recognition}.

\bibitem[{Krojer et~al.(2022)Krojer, Adlakha, Vineet, Goyal, Ponti, and
  Reddy}]{Krojer:etal:2022}
Benno Krojer, Vaibhav Adlakha, Vibhav Vineet, Yash Goyal, Edoardo Ponti, and
  Siva Reddy. 2022.
\newblock Image retrieval from contextual descriptions.
\newblock In \emph{Proceedings of ACL}, pages 3426--3440, Dublin, Ireland.

\bibitem[{Lake et~al.(2017)Lake, Ullman, Tenenbaum, and
  Gershman}]{Lake:etal:2016}
Brenden Lake, Tomer Ullman, Joshua Tenenbaum, and Samuel Gershman. 2017.
\newblock Building machines that learn and think like people.
\newblock \emph{Behavorial and Brain Sciences}, 40:1--72.

\bibitem[{Li et~al.(2019)Li, Wong, Zhao, and Kankanhalli}]{Li:etal:2019}
Junnan Li, Yongkang Wong, Qi~Zhao, and Mohan Kankanhalli. 2019.
\newblock Video storytelling: {Textual} summaries for events.
\newblock \emph{IEEE Transactions on Multimedia}, 22(2):554--565.

\bibitem[{Lin et~al.(2014)Lin, Maire, Belongie, Hays, Perona, Ramanan,
  Doll{\'a}r, and Zitnick}]{Lin:etal:2014}
Tsung-Yi Lin, Michael Maire, Serge Belongie, James Hays, Pietro Perona, Deva
  Ramanan, Piotr Doll{\'a}r, and C.~Lawrence Zitnick. 2014.
\newblock Microsoft {COCO}: {Common} objects in context.
\newblock In \emph{Computer Vision -- ECCV 2014}.

\bibitem[{Lu et~al.(2019)Lu, Batra, Parikh, and Lee}]{Lu:etal:2019}
Jiasen Lu, Dhruv Batra, Devi Parikh, and Stefan Lee. 2019.
\newblock {ViLBERT}: {Pretraining} task-agnostic visiolinguistic
  representations for vision-and-language tasks.
\newblock In \emph{Proceedings of NeurIPS}, Vancouver, Canada.
\newblock Published online: \url{https://papers.nips.cc/paper/2019}.

\bibitem[{Lu et~al.(2022)Lu, Bartolo, Moore, Riedel, and
  Stenetorp}]{Lu:etal:2022}
Yao Lu, Max Bartolo, Alastair Moore, Sebastian Riedel, and Pontus Stenetorp.
  2022.
\newblock Fantastically ordered prompts and where to find them: {O}vercoming
  few-shot prompt order sensitivity.
\newblock In \emph{Proceedings of ACL}, pages 8086--8098, Dublin, Ireland.

\bibitem[{Mokady et~al.(2021)Mokady, Hertz, and Bermano}]{Mokady:etal:2021}
Ron Mokady, Amir Hertz, and Amit Bermano. 2021.
\newblock {ClipCap}: {CLIP} prefix for image captioning.
\newblock \url{https://arxiv.org/abs/2111.09734}.

\bibitem[{Peirce et~al.(2019)Peirce, Gray, Simpson, MacAskill,
  H\"{o}chenberger, Sogo, Kastman, and Lindel{\o}v}]{Peirce2019}
Jonathan Peirce, Jeremy~R. Gray, Sol Simpson, Michael MacAskill, Richard
  H\"{o}chenberger, Hiroyuki Sogo, Erik Kastman, and Jonas~Kristoffer
  Lindel{\o}v. 2019.
\newblock \href {https://doi.org/10.3758/s13428-018-01193-y} {{PsychoPy}2:
  Experiments in behavior made easy}.
\newblock \emph{Behavior Research Methods}, 51(1):195--203.

\bibitem[{Radford et~al.(2021)Radford, Kim, Hallacy, Ramesh, Goh, Agarwal,
  Sastry, Askell, Mishkin, Clark, Krueger, and Sutskever}]{Radford:etal:2021}
Alec Radford, Jong~Wook Kim, Chris Hallacy, Aditya Ramesh, Gabriel Goh,
  Sandhini Agarwal, Girish Sastry, Amanda Askell, Pamela Mishkin, Jack Clark,
  Gretchen Krueger, and Ilya Sutskever. 2021.
\newblock Learning transferable visual models from natural language
  supervision.
\newblock In \emph{Proceedings of ICML}, pages 8748--8763, virtual conference.

\bibitem[{Radford et~al.(2019)Radford, Wu, Child, Luan, Amodei, and
  Sutskever}]{Radford:etal:2019}
Alec Radford, Jeffrey Wu, Rewon Child, David Luan, Dario Amodei, and Ilya
  Sutskever. 2019.
\newblock Language models are unsupervised multitask learners.
\newblock
  \url{https://d4mucfpksywv.cloudfront.net/better-language-models/language-models.pdf}.

\bibitem[{Rohrbach et~al.(2018)Rohrbach, Hendricks, Burns, Darrell, and
  Saenko}]{Rohrbach:etal:2018}
Anna Rohrbach, Lisa~Anne Hendricks, Kaylee Burns, Trevor Darrell, and Kate
  Saenko. 2018.
\newblock Object hallucination in image captioning.
\newblock In \emph{Proceedings of EMNLP}, pages 4035--4045, Brussels, Belgium.

\bibitem[{Shin et~al.(2020)Shin, Razeghi, Logan~IV, Wallace, and
  Singh}]{Shin:etal:2020}
Taylor Shin, Yasaman Razeghi, Robert Logan~IV, Eric Wallace, and Sameer Singh.
  2020.
\newblock {A}uto{P}rompt: {E}liciting knowledge from language models with
  automatically generated prompts.
\newblock In \emph{Proceedings of EMNLP}, pages 4222--4235, virtual conference.

\bibitem[{Wallace et~al.(2019)Wallace, Feng, Kandpal, Gardner, and
  Singh}]{Wallace:etal:2019}
Eric Wallace, Shi Feng, Nikhil Kandpal, Matt Gardner, and Sameer Singh. 2019.
\newblock Universal adversarial triggers for attacking and analyzing {NLP}.
\newblock In \emph{Proceedings of EMNLP}, pages 2153--2162, Hong Kong, China.

\bibitem[{Webson and Pavlick(2022)}]{Webson:Pavlick:2022}
Albert Webson and Ellie Pavlick. 2022.
\newblock Do prompt-based models really understand the meaning of their
  prompts?
\newblock In \emph{Proceedings of NAACL}, pages 2300--2344, Seattle, WA.

\bibitem[{Xu et~al.(2016)Xu, Mei, Yao, and Rui}]{Xu:etal:2016}
Jun Xu, Tao Mei, Ting Yao, and Yong Rui. 2016.
\newblock {MSR-VTT}: {A} large video description dataset for bridging video and
  language.
\newblock In \emph{Proceedings of CVPR}, pages 5288--5296, Sydney, Australia.

\bibitem[{Zeng et~al.(2022)Zeng, Attarian, Ichter, Choromanski, Wong, Welker,
  Tombari, Purohit, Ryoo, Sindhwani, Lee, Vanhoucke, and
  Florence}]{Zeng:etal:2022}
Andy Zeng, Maria Attarian, Brian Ichter, Krzysztof Choromanski, Adrian Wong,
  Stefan Welker, Federico Tombari, Aveek Purohit, Michael Ryoo, Vikas
  Sindhwani, Johnny Lee, Vincent Vanhoucke, and Pete Florence. 2022.
\newblock Socratic models: {Composing} zero-shot multimodal reasoning with
  language.
\newblock \url{https://arxiv.org/abs/2204.00598}.

\end{thebibliography}
\bibliographystyle{acl_natbib}

\appendix

\section{Neural caption generation details}
\label{appendix:model-details}
In our experiments, we use the pre-trained ClipCap caption generation model from \citet{Mokady:etal:2021}, which employs a Transformer mapper trained on the COCO data-set \cite{Lin:etal:2014} while the CLIP image encoder \citep{Radford:etal:2021} and the GPT-2 language model \citep{Radford:etal:2019} are frozen. 
In the ClipCap architecture, the mapper projects a CLIP-extracted embedding into the multidimensional space of GPT-2 word embeddings to trigger image-conditioned text generation. ClipCap has two architectural variants, one that uses a Transformer-based mapper and one that employs an MLP-based mapper. We refer to \citet{Mokady:etal:2021} for a detailed description of the MLP variant. The model that uses a Transformer mapper extracts a visual embedding from a pre-trained CLIP image encoder and feeds such representation together with a set of learned constant embeddings into GPT-2.

To produce a caption, we generate text using beam search with 5 beams, without tuning this value, and retaining the single maximum likelihood sequence. We set a maximum caption length of 67 tokens. Given that neural captions have an average length of around 11 tokens, it is unlikely that this limit is of any practical import. Additionally, the pre-trained CLIP text encoder from \citet{Radford:etal:2021}, which both \citet{Krojer:etal:2022} and we use, cannot process contexts larger than 75 tokens, and thus extra tokens would be ignored in any case.

\section{Caption frequency distribution analysis}
\label{appendix:frequency-analysis}

We tokenize, part-of-speech tag and lemmatize human and neural captions with Spacy.\footnote{\url{https://spacy.io/}} We use the resulting part-of-speech and lemma sequences to compute the statistics reported in this Appendix and in Appendix \ref{appendix:typical-words}.

\begin{figure}[tb]
    \centering
    \includegraphics[width = 0.45\textwidth]{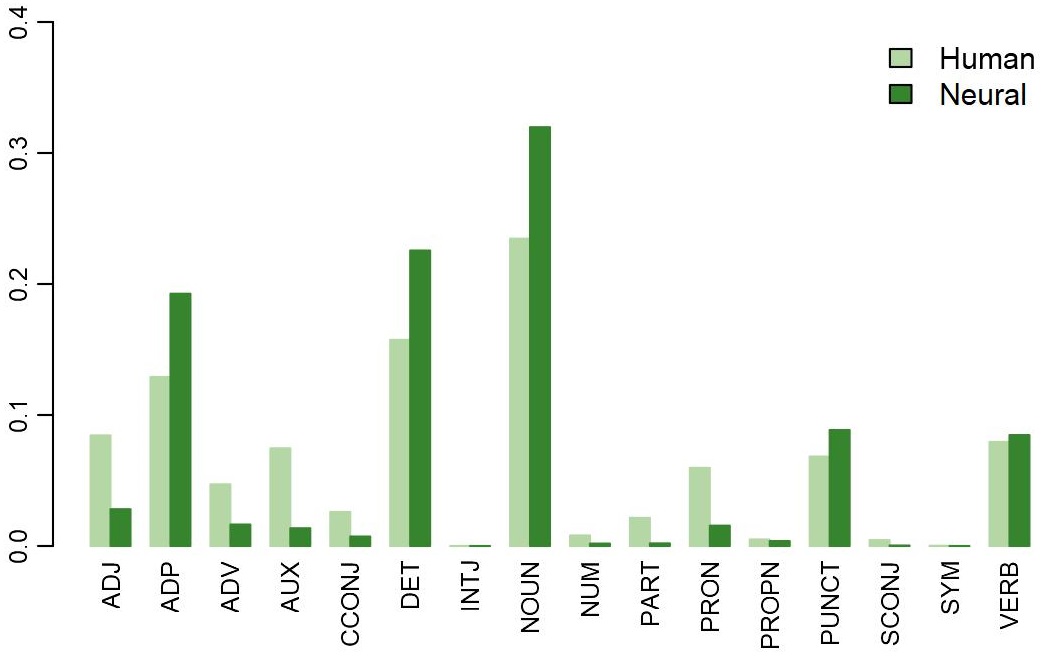}
    \caption{Part of Speech frequency distribution in human and neural captions.}
    \label{fig:pos_distribution}
\end{figure}

We counted the occurrences of the different parts of speech, normalized over the total amount of produced tokens, in both human and neural captions (Fig.~\ref{fig:pos_distribution}). 
Their distribution reveals that, unsurprisingly, both types of caption mostly use nouns to denote the entities presented in the images, but humans tend to modify them with remarkably higher adjective usage. Human-generated captions also display more functional words, pointing to the higher syntactic complexity already suggested by sentence length. 

\begin{figure}[tb]
    \centering
    \includegraphics[width = 0.45\textwidth]{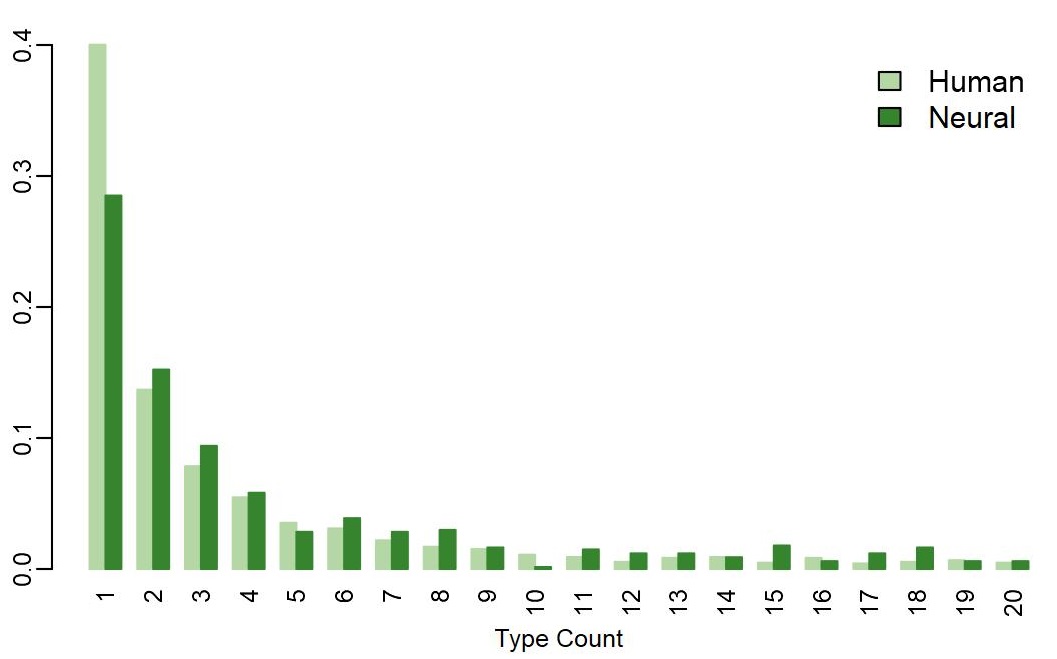}
   \caption{Lemma frequency spectrum in human and neural captions (only frequency of first 20 counts shown). The x-axis represents an occurrence count, the y-axis the number of distinct lemmas with that count in the captions, normalized over the total amount of distinct lemma types.}
    \label{fig:freq_spectrum_lemma}
\end{figure}

We computed the frequency spectrum \citep{Baayen:2001} of lemma types occurring in the two sets of captions. The normalized count of distinct lemmas with caption occurrence from 1 to 20 are plotted in Fig.~\ref{fig:freq_spectrum_lemma}. Human captions make a larger use of lemmas occurring only once, displaying a clear Zipfian trend. This trend is also present, but much less pronounced, in the neural captions.

\section{Neural vs.~human caption lemma analysis}
\label{appendix:typical-words}

We use the Local Mutual Information score \citep{Evert:2005} to extract lemmas that are most significantly associated with neural vs.~human captions, based on their relative frequency of occurrence in the two sets.
\\
\\
\noindent{}\textbf{Top 20 most typical lemmas of neural caption set} (min LMI: 124.2): \textit{a, stand, of, in, next, hold, people, woman, group, front, on, sit, man, person, table, with, food, couple, cell, phone.}
\\
\\
\noindent{}\textbf{Top 20 most typical lemmas of human caption set} (min LMI: 89.7): \textit{the, be, right, see, left, can, and, 's, visible, you, her, have, hand, his, not, there, at, face, but, just.}
\\
\\
Besides looking at lemmas most typical of each set, we explore whether there is some non-trivial degree of overlap between the words occurring in the neural vs.~human captions for each target (excluding stop words and punctuation, that would artificially increase overlap). We find that the average lemma overlap, measured as intersection-over-union (IOU), is at 5.2\% (st.~dev.: 6.6\%). This might look non-negligible, but it does not significantly differ from random overlap according to a permutation test. %
% However, considering the large variance and the fact that the average IOU across 100 random shufflings of the captions (so that neural and human captions are no longer matched by target) is at 1.5\%, with 3.8\% standard deviation, we deem the observed degree of IOU not significantly above what expected by chance.

\section{Crowdsourcing experiment details}
\label{appendix:crowdsourcing}

\begin{figure*}[tb]
  \centering
  \subfloat[]{\includegraphics[width=0.49\textwidth]{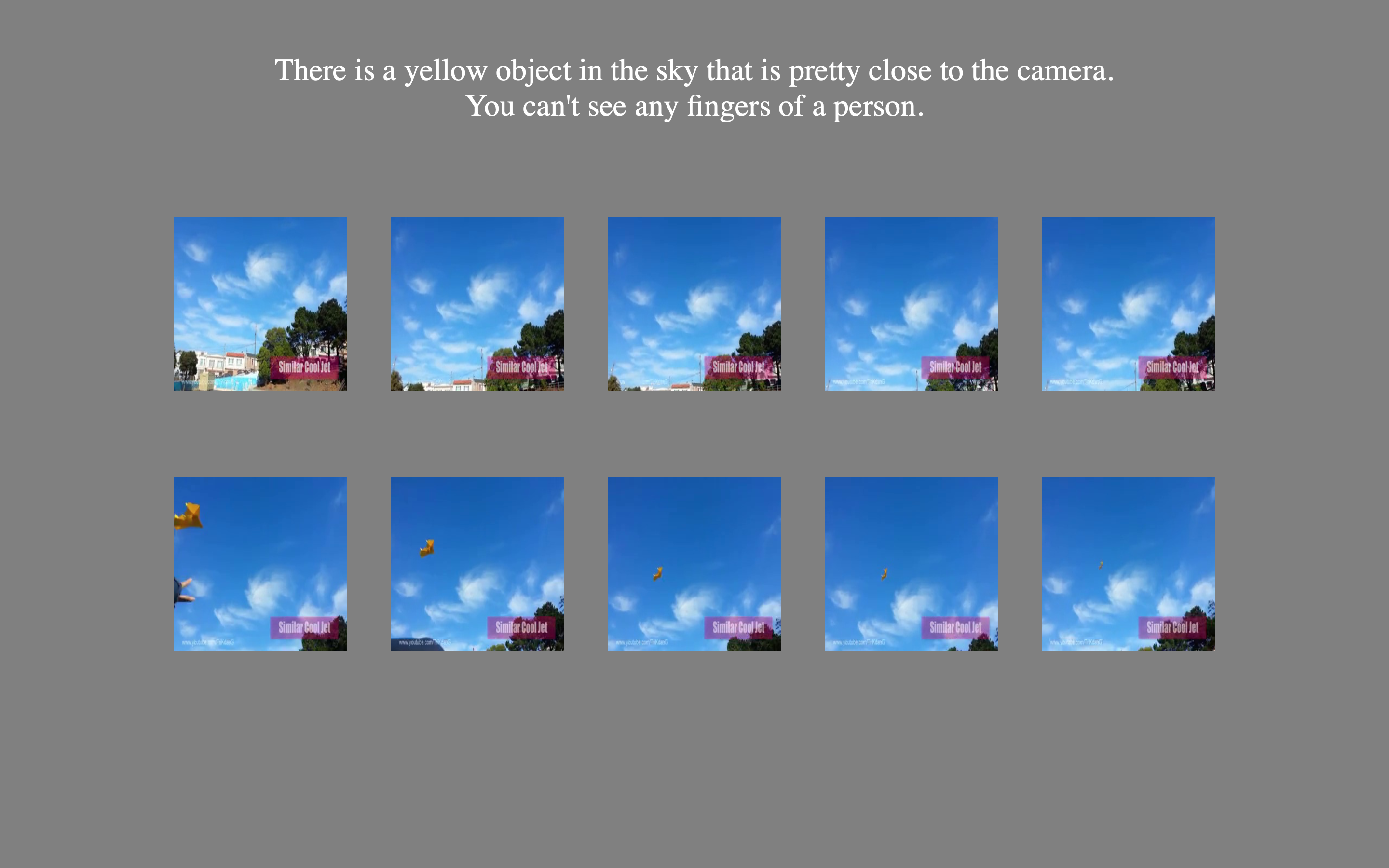}\label{fig:example_annotation}}
  \hfill
  \subfloat[]{\includegraphics[width=0.49\textwidth]{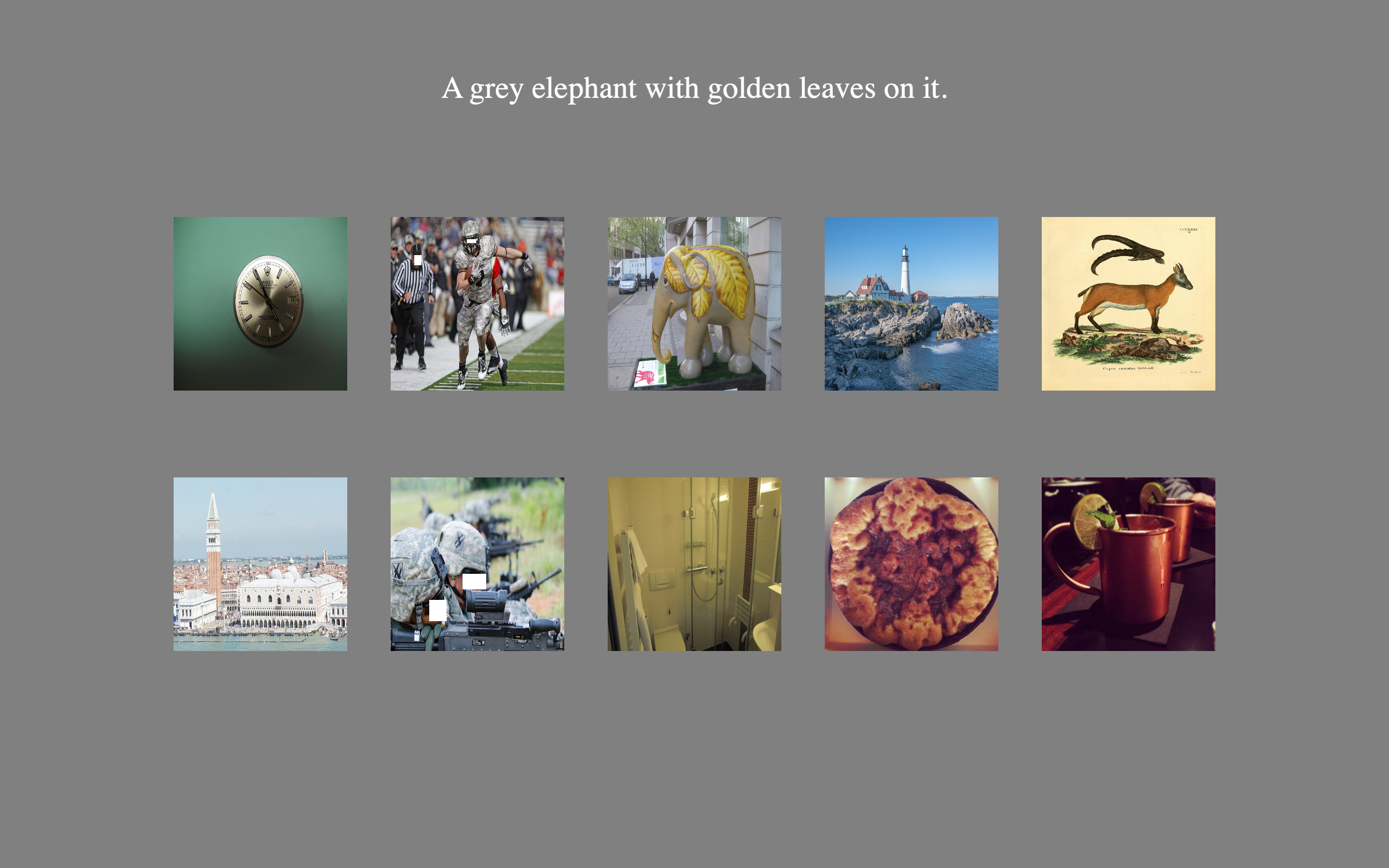}\label{fig:example_check}}
  \caption{Examples of screens shown to the participants. In panel (a), a set with a human caption; In panel (b), an attention check.}
\end{figure*}

We populated the \IC{} stimulus subset for the human retrieval experiment as follows. We took all 522 candidate sets where the neural retriever guessed the target from the \IC{} \textit{video} section. We further sampled without replacement the same amount of cases from the sets that the retriever got wrong (thus obtaining a balanced sample where the neural retriever accuracy is at 50\%). We presented to subjects these 1,044 sets with both the human captions from \citet{Krojer:etal:2022} and the captions produced by our neural caption generation system. This resulted in 2,088 questions posed to subjects.

We randomly divided the entire set into 36 blocks of 58 questions (always containing both neural and human captions in similar amounts).  In each screen, the 10 images from a set were presented at the center, arranged in two arrays of 5 images, with the caption written above--see Fig.~\ref{fig:example_annotation}. Participants were asked to click on the image that matched the caption best. They were shown one example before starting the task. They were also warned that some cases could be more challenging than others. We asked them to always reply with the answer they found most plausible. Finally, they were warned that the experiment contained some control items, used to ensure annotation quality.

Each subject was presented with one block of questions, plus 5 randomly placed controls, designed to ensure that annotators were paying attention to the task. These cases were made intentionally very simple: targets were surrounded by random distractors, i.e., images that were neither contextually relevant nor very similar to the target. Targets and distractors for the attention checks were extracted from the less challenging \textit{static} section of the \IC{} data-set \citep{Krojer:etal:2022}. We made sure internally that these sets could be easily processed with 100\% retrieval accuracy. See Fig.~\ref{fig:example_check} for an example. 

The data collection routine was written in Psychopy \citep{Peirce2019} and launched through Pavlovia.\footnote{\url{https://pavlovia.org/}} There was no time limit for completing the study. 

We recruited participants via Amazon Mechanical Turk.\footnote{\url{https://www.mturk.com/}} We only accepted annotators from the US, with HIT approval rate higher than 89\% and number of approved HITs higher than 1,000. We informed them that we would not collect any personal data (except for their workerID, that we would not make public), and that the goal of the experiment was to study how well people identify images based on descriptions. Before being able to access the link of the experiment, participants had to complete an informed consent form. They were able to quit the experiment at any time. We paid them 12.5\$ for completing the task. The experiment was approved by the ethical board of Universitat Pompeu Fabra in the context of the AMORE project (grant agreement No.\ 715154).

\begin{figure}[tb]
  \centering
  \includegraphics[width=\columnwidth]{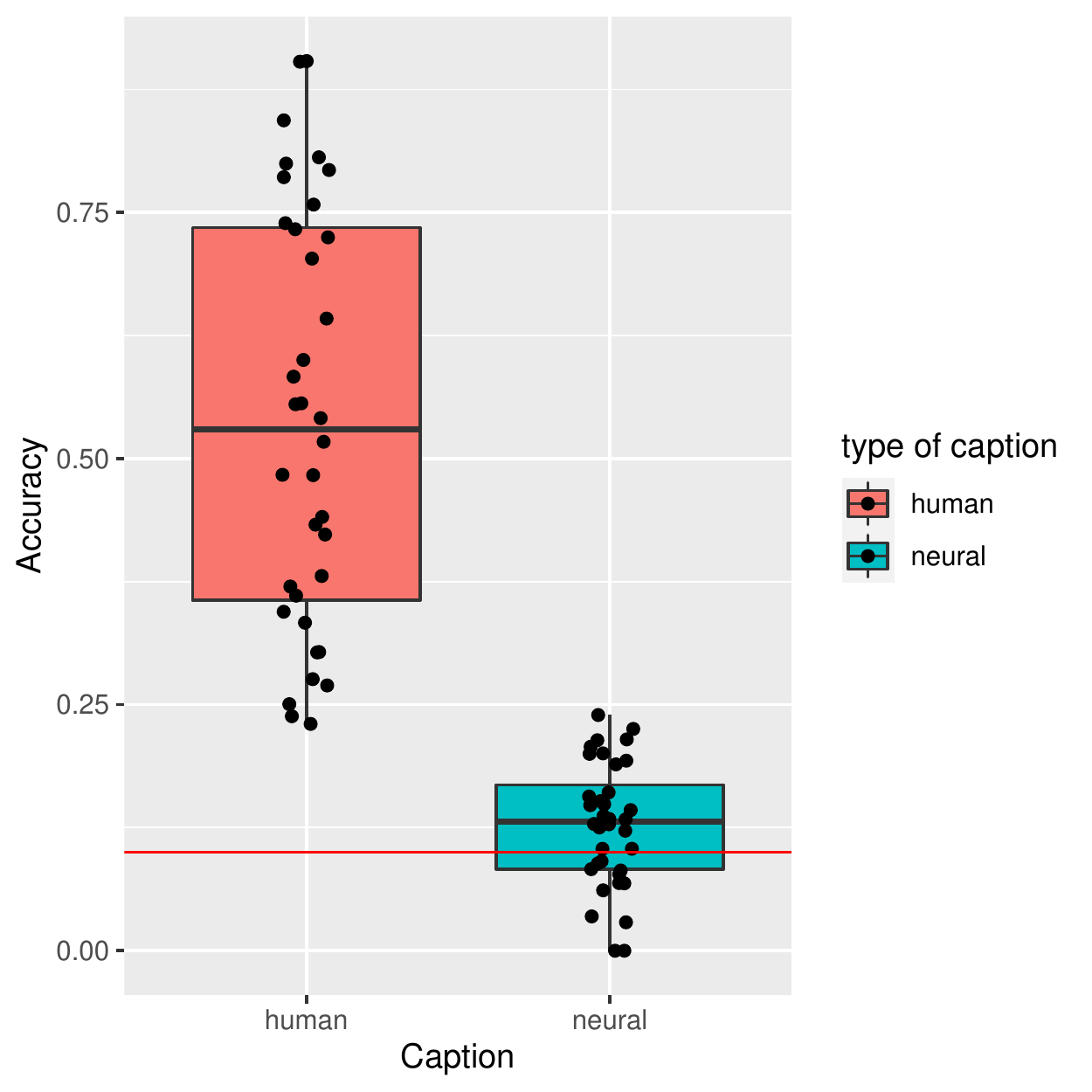}
  \caption{Accuracy of our 36 participants on human vs.~neural captions. Each point represents one participant. The red line represents chance level.}
  \label{fig:boxplot}
\end{figure}

We excluded the data of participants that made more than one mistake when scoring the controls, suggesting that they were not paying enough attention to the task. After a first round of data collection, we computed mean accuracy and standard deviation on human captions (without looking at neural caption performance). To further filter out low-quality trials, we removed participants with human caption accuracy more than one standard deviation below the mean, again suggesting scarce attention to the task. This resulted in 6 participants being removed, with the corresponding data being collected again. The boxplot in Fig.~\ref{fig:boxplot} shows the distribution of accuracy on human and neural captions for our final 36 participants.  All participants reached well-above-chance accuracy on human captions, with a clear contrast with respect to their neural caption performance (the worst performance on human captions is comparable to the best performance on neural captions).

Our final cumulative human accuracy on human captions is considerably lower than the one reported by \citet{Krojer:etal:2022} for the whole \IC{} collection (54.3\% vs.~90\%). We conjecture that this is due in part to the fact that our items only come from the more challenging \IC{} \textit{video} subset, and in part to the fact that their two-stage data-collection setup allowed a subject white-listing procedure we could not implement. Still, three authors performed the caption retrieval task, with resulting accuracies at 52\%, 56\% and 76\%, respectively. The performance distribution of this supposedly ``high-quality'' annotators is comparable to the one of the 36 crowd-sourced participants, suggesting that the low overall accuracy is genuinely due to the difficulty of the task, and not to poor quality control.

\section{Fig.~\ref{fig:SixExamples} example selection method and full candidate sets}
\label{appendix:example-figure-construction}
The examples in Fig.~\ref{fig:SixExamples} were randomly selected among trials in which, given the neural caption, the neural model guessed the target and the human annotators missed it. We avoided re-sampling the same candidate set more than once. We also discarded images displaying identifiable persons or large portions of text. 

In each example, the target image is presented with a distractor, which can be the frame immediately preceding the target or the frame following it in the original sequence. The choice to show the preceding vs.~following distractor frame was random.

\begin{figure*}[p]
    \centering
    \includegraphics[scale = 0.70]{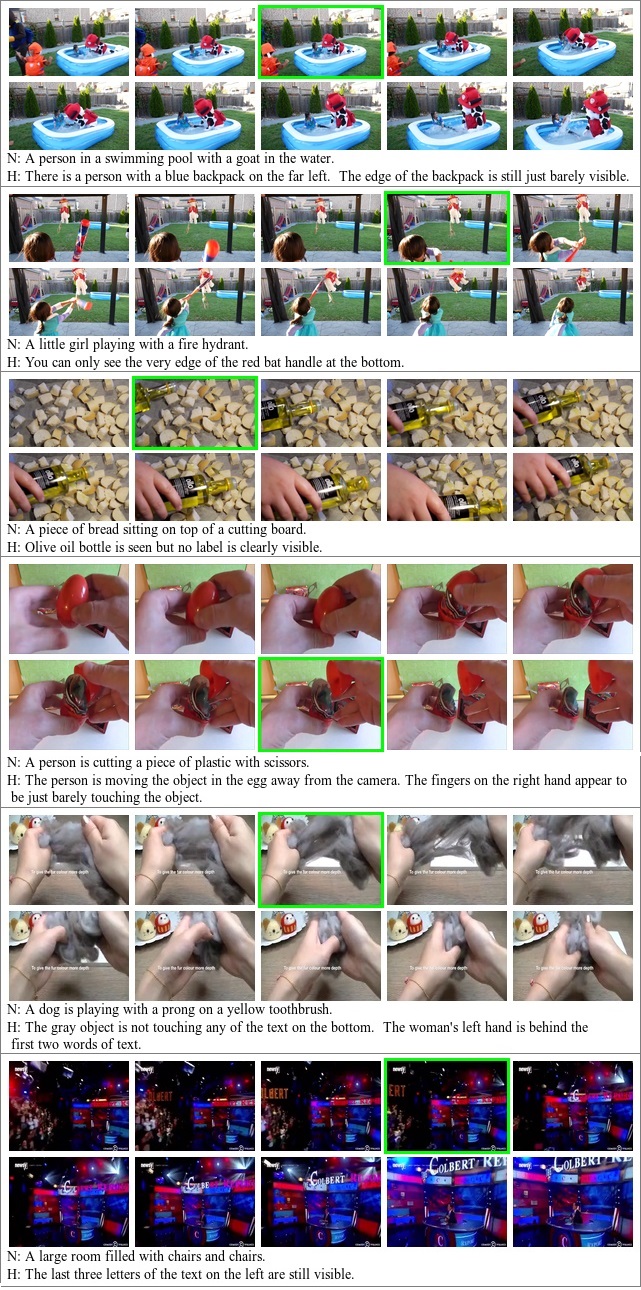}
    \caption{Whole candidate sets for each example in Fig.~\ref{fig:SixExamples}. The target image is marked in green.}
    \label{fig:fullsets}
\end{figure*}

In Fig.~\ref{fig:fullsets}, we report the full sequences of distractors of each selected set, with the target marked in green, and the corresponding captions produced by the neural model (N), and by humans (H). 

\end{document}